\documentclass[final]{cvpr}

\usepackage{multirow}
\usepackage{tablefootnote}

\usepackage{times}
\usepackage{epsfig}
\usepackage{graphicx}
\usepackage{amsmath}
\usepackage{amssymb}
\usepackage{soul}
\usepackage[dvipsnames]{xcolor}

\usepackage[pagebackref=true,breaklinks=true,colorlinks,bookmarks=false]{hyperref}

\def\cvprPaperID{5671} 
\def\confYear{CVPR 2021}

\begin{document}

\title{Spatially Consistent Representation Learning}

\author{Byungseok Roh\thanks{Equal contribution} \qquad Wuhyun Shin\footnotemark[1] \qquad Ildoo Kim \qquad Sungwoong Kim\\
Kakao Brain\\
{\tt\small \{peter.roh, aiden.hsin, ildoo.kim, swkim\}@kakaobrain.com}
}

\maketitle

\begin{abstract}

Self-supervised learning has been widely used to obtain transferrable representations from unlabeled images. Especially, recent contrastive learning methods have shown impressive performances on downstream image classification tasks. While these contrastive methods mainly focus on generating invariant global representations at the image-level under semantic-preserving transformations, they are prone to overlook spatial consistency of local representations and therefore have a limitation in pretraining for localization tasks such as object detection and instance segmentation. Moreover, aggressively cropped views used in existing contrastive methods can minimize representation distances between the semantically different regions of a single image. 

In this paper, we propose a spatially consistent representation learning algorithm (\textit{SCRL}) for multi-object and location-specific tasks. In particular, we devise a novel self-supervised objective that tries to produce coherent spatial representations of a randomly cropped local region according to geometric translations and zooming operations. On various downstream localization tasks with benchmark datasets, the proposed SCRL shows significant performance improvements over the image-level supervised pretraining as well as the state-of-the-art self-supervised learning methods.    
Code is available at \url{https://github.com/kakaobrain/scrl}.
    
\end{abstract}

\begin{figure}[t]
\centering
\def\arraystretch{1.0}
\setlength{\tabcolsep}{0.2em}
\begin{tabular}{cc}
\includegraphics[trim={0 0 21.5cm 0},clip,height=0.79\linewidth]{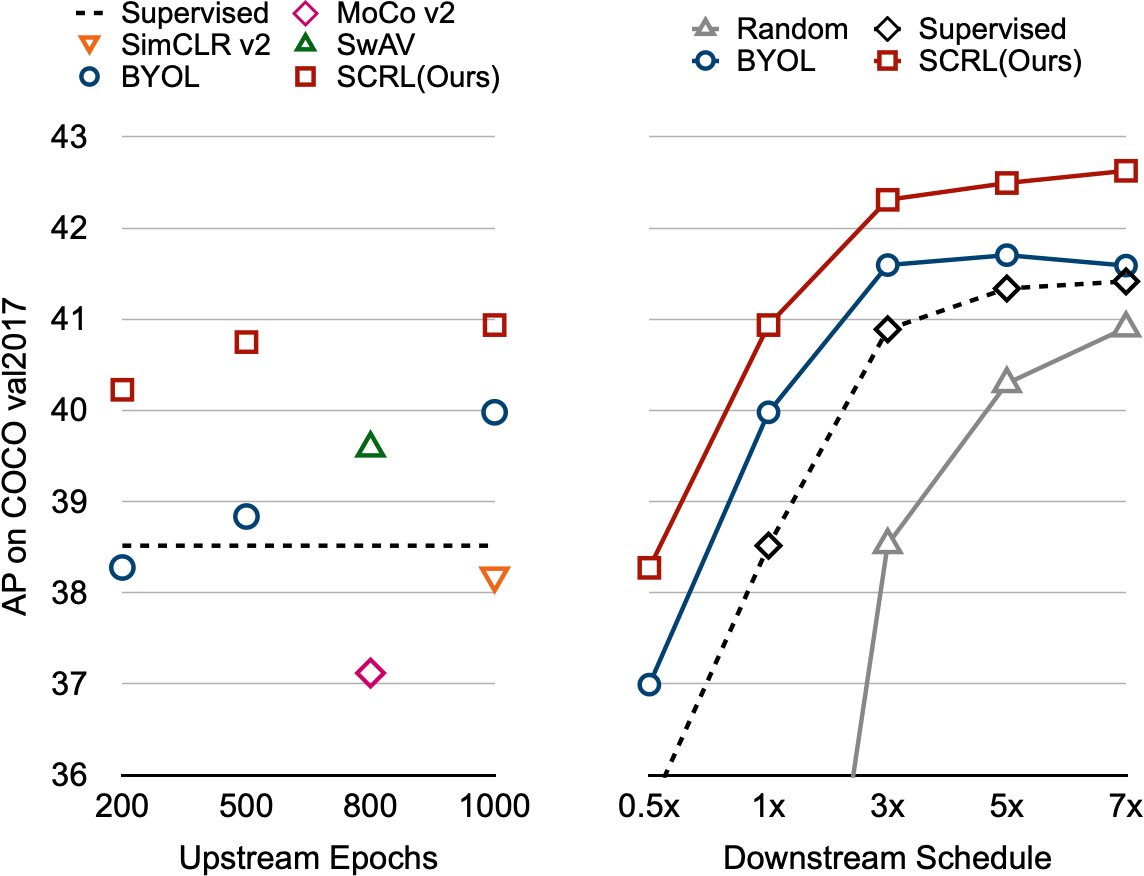} &
\includegraphics[trim={21.5cm 0 0 0},clip,height=0.79\linewidth]{figures/schedule_up_down.png} \\
{\footnotesize (a) } & {\footnotesize (b)} 
\end{tabular}
\vspace{0mm}
\caption{\textbf{(a)} AP on downstream of COCO detection task \wrt the upstream epochs on ImageNet. We use a ResNet-50-FPN backbone with Faster R-CNN, using default training configuration used in \cite{detectron2}. Only with 200 epochs of upstream, SCRL outperforms the ImageNet pre-trained counterpart as well as the state-of-the-art self-supervised learning methods. \textbf{(b)} AP on COCO detection task under varied downstream schedules from 0.5$\times$ (45k iterations) to 7$\times$ (630k iterations). SCRL consistently outperforms random initialization, supervised pretraining, and BYOL in all the training schedules.}
\label{fig:coco_results}
\end{figure}

\section{Introduction}

In computer vision, unsupervised representation learning from a large amount of unlabeled images has been shown to be effective in improving the performances of neural networks for unknown downstream tasks, especially with few labeled data \cite{Carl2017iccv, Xiao2020}. While conventional generative modeling algorithms are difficult to obtain semantically meaningful representations from high-resolution natural images due to their focus on low-level details \cite{BIGBIGAN, IGPT}, self-supervised learning algorithms have recently shown promising results in obtaining semantic representations via the use of proxy tasks on unsupervised data \cite{Exemplar, JigsawPuzzle, Colorization, Rotation, PIRL, MoCov1, SimCLRv1, InfoNCE, SwAV, BYOL}. Among them, contrastive learning methods with discriminative models have particularly achieved remarkable performances on most downstream tasks related to image classification problems \cite{PIRL, MoCov1, MoCov2, SimCLRv1, SimCLRv2, InfoNCE, SwAV, BYOL, CPCv2, InfoMin}. 

Contrastive self-supervised learning aims to obtain discriminative representations based on the semantically positive and negative image pairs. Specifically, it tries to produce invariant representations from semantic-preserving augmentations of the same image while making representations dissimilar from different images. However, most existing contrastive methods exploit consistent global representations on a per image basis, specific for image classification, and therefore they are likely to generate inconsistent local representations with respect to the same spatial regions after image transformations. For example, when a certain object in an image is geometrically shifted or scaled, previous global contrastive methods can produce a similar global representation, even if the local feature of that object ends up losing consistency\cite{SNIP}, since they use global pooling by which they can attend to other discriminative areas instead. This can consequently lead to performance degradation on localization tasks based on spatial representations. In addition, previous contrastive methods often utilize heavily cropped views from an image to make a positive pair, and hence the representations between the semantically different regions are rather induced to be matched \cite{Demystifying}.

In order to resolve these issues on the existing global contrastive learning methods, we propose a spatially consistent representation learning algorithm, \textit{SCRL}, that can leverage lots of unlabeled images, specifically for multi-object and location-specific downstream tasks including object detection and instance segmentation. In specific, we develop a new self-supervised objective to realize the invariant spatial representation corresponding to the same cropped region under augmentations of a given image. 
Since we are able to figure out the two exactly matched spatial locations for each cropped region on the two transformed images, each positive pair of cropped regions necessarily has a common semantic information. From a positive pair of cropped feature maps, we apply RoIAlign \cite{MaskRCNN} to the respective maps and obtain equally-sized local representations. We optimize the encoding network to minimize the distance between these two local representations. Since BYOL \cite{BYOL} has shown to be an efficient contrastive learning method without requiring negative pairs, we adapt its learning framework for producing our spatially coherent representations.

We perform extensive experiments and analysis on several benchmark datasets to empirically demonstrate the effectiveness of the proposed SCRL in significantly improving the performances of fine-tuned models on various downstream localization tasks. Namely, SCRL consistently outperforms the random initialization, the previous image-level supervised pretraining and the state-of-the-art self-supervised methods, on the tasks of object detection and instance segmentation, with the PASCAL VOC, COCO and Cityscapes datasets. In particular, SCRL leads to regress object boundaries more precisely owing to accurate spatial representations before being fed into the task-specific head networks. Importantly, as shown in Figure \ref{fig:coco_results}, SCRL outperforms the other pretraining methods even with a small number of epochs during upstream training on unlabeled images. In addition, the improvements in fine-tuned downstream performance obtained by SCRL are consistently maintained under longer schedules as well as small data regime(\ie, 1/10 of COCO training data), which validates the benefits of transferred spatial representations by SCRL.

\vspace{2mm}
Our main contributions can be summarized as follows:
\begin{itemize}
\item We take into account spatial consistency rather than global consistency on image representations and propose a novel self-supervised learning algorithm, SCRL, on unlabeled images, especially for multi-object and location-aware downstream tasks.  
\item We generate multiple diverse pairs of semantically-consistent cropped spatial feature maps and apply an efficient contrastive learning method with a dedicated local pooling and projection. 
\item A variety of experimental results show clear advantages of SCRL over the existing state-of-the-art methods as a transferrable representation pretraining in obtaining better performances on localization tasks.
\end{itemize}


\vspace{2mm}
\section{Related Work}

Early approaches for self-supervised learning rely on hand-crafted proxy tasks \cite{Exemplar, JigsawPuzzle, Colorization, Rotation} from which the models can extract meaningful information that are beneficial to the considered downstream tasks.
However, the representation obtained by those works are prone to lose generality due to the strong prior knowledge reflected to the design choice of pretext tasks.

Recently, contrastive methods \cite{PIRL, MoCov1, MoCov2, SimCLRv1, SimCLRv2} have made a lot of progress in the field of self-supervised learning. The goal of contrastive learning is to minimize the distances between the positive pairs, namely, two different augmented views of a single image. At the same time, negative pairs should be pushed apart, which can be directly encouraged by training objectives, such as InfoNCE \cite{InfoNCE}. PIRL \cite{PIRL} tries to learn invariant features under semantic-preserving transformations. MoCo \cite{MoCov1, MoCov2} focuses on constructing a minibatch with a large number of negative samples by utilizing the dynamic queuing and the moving-averaged encoder. SimCLR \cite{SimCLRv1, SimCLRv2} improves the quality of representation by finding a more proper composition of transformations and an adequate size of non-linear heads at the top of the network. Those methods, however, generally require a larger batch size compared to the supervised counterpart in order to avoid mode collapse problems. 

\begin{figure*}[t]
\centering
\includegraphics[width=1.0\linewidth]{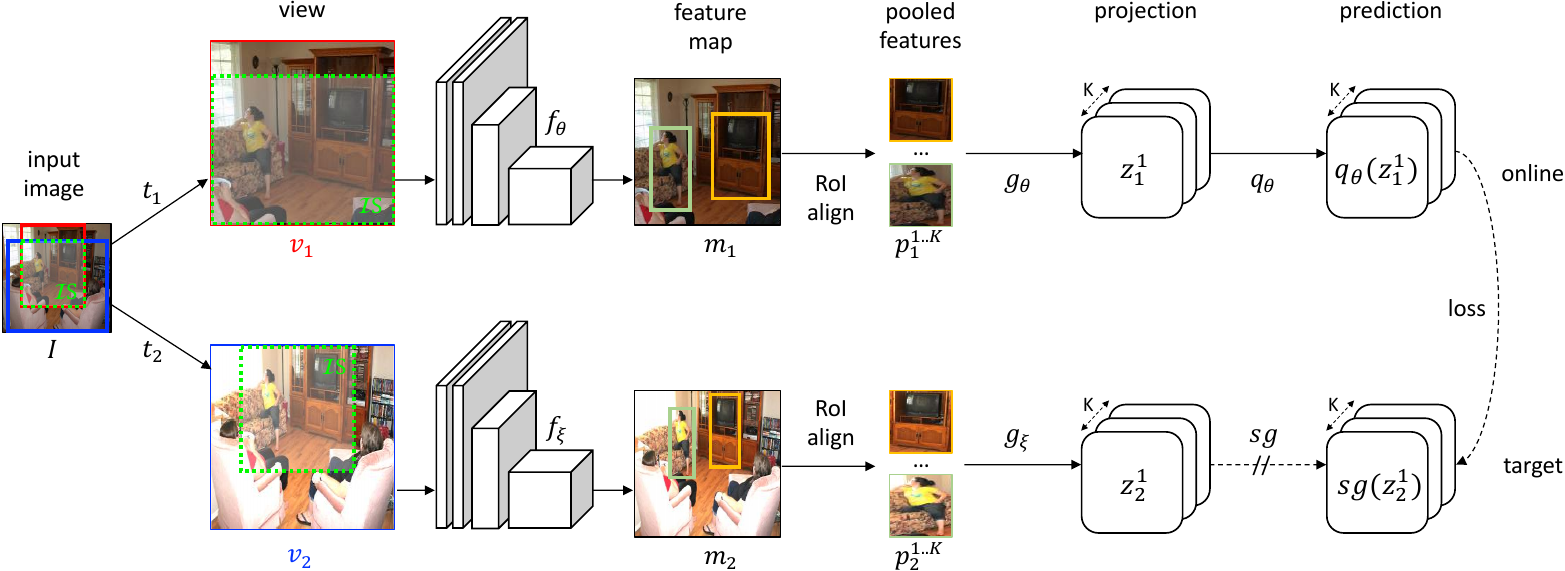}
\caption{An illustrative view of our method. We first find the intersection region {\color{green}$\mathcal{IS}$} between {\color{red}$v_1$} and {\color{blue}$v_2$} to randomly generate $K$ number of RoIs within {\color{green}$\mathcal{IS}$}. SCRL minimizes a similarity loss between the predictions of the pooled RoIs $q_{\theta}(z_{1}^{k})$ for $v_1$ and the projections of the pooled RoIs $sg(z_{2}^{k})$ for $v_2$, where the online network's parameters $\theta$ are trained, parameters of the target network $\xi$ are updated by an exponential moving average of $\theta$, and $sg$ stands for stop-gradient. At the end of training, everything but $f_{\theta}$ is discarded. (Images located in `feature map' and `pooled features' are not real features, we use $v_1$ and $v_2$ images for a better understanding.) }
\vspace{-3mm}
\label{fig:scrl_architecture}
\end{figure*}

More recently, SwAV \cite{SwAV} modifies previous pairwise representation comparisons by introducing cluster assignment and swapped prediction. They also propose a novel augmentation strategy, multi-crop, which appears to be similar to our SCRL in that they compare multiple smaller patches cropped from the same image, but substantially different in that the spatial consistency on the feature map is not directly considered. Grill \etal \cite{BYOL} devises the method named BYOL where the network bootstraps its own representation by keeping up with the moving averaged version of itself. BYOL removes the necessity of negative pairs and have shown to be more robust against changes in batch size. However, they also still overlook the object-level local consistency, and moreover, there always exists, as pointed out in \cite{Demystifying}, a chance that aggressive views taken from a single image can have different semantic meanings, especially in the object-level. 

There also has been a large body of works that leverage geometric correspondence for learning dense representation. While the early efforts \mbox{\cite{MatchNet, ComparePatch, WarpNet, DeepFeature, LFNet}} mostly rely on explicit supervisory signals, some of recent works adopt self-supervised methods to learn the parts or landmarks of the data. Similar to our work, Thewlis \mbox{\etal} \mbox{\cite{EquivariantImageLabelling}} employ siamese framework to match label maps from the different views of an image by making use of equivariance relation between them. DVE {\cite{DVE}} extends this work to even take into account the correspondence across different instances that shares the same object category. In this vein of works, their interests are only limited to dense representation itself that comes with object structure learning, whereas our work aims to transfer the learned representation to a wide variety of localization downstream tasks.

Concurrent to our work, VADeR {\cite{VADeR}} also similarly learns pixel-level representation in order to transfer it to multiple dense prediction tasks. While VADeR constructs the positive pairs from a discrete set of pixels, our method, SCRL, can possibly sample infinite number of pairs by pooling the variable-sized  random regions with bilinear interpolation. This means that VADeR can be viewed as a specific instance of our method where the sizes of the pooled box is fixed to a single pixel without a sophisticated pooling technique. Furthermore, VADeR utilizes an extra decoder architecture, while SCRL exploits an encoder-only structure.


\section{Method}
\label{sec:method}

This section describes the proposed SCRL for downstream localization tasks in detail. We first present the proposed pair of locally cropped boxes to be matched from different views of the same image. Then, the proposed self-supervised objective based on the spatial consistency loss is defined. The details in implementation of our self-supervised learning are finally presented.

\subsection{Spatially Consistent Representation Learning}

Motivated by BYOL \cite{BYOL}, we use two neural networks: the \emph{online network} which is defined by a set of parameters $\theta$ and \emph{target network} parameterized by $\xi$. 
The target network provides the regression target to train the online network while the target network's parameter set $\xi$ follows the online network's parameter set $\theta$ by using an exponential moving average with a decay parameter $\tau$, \ie, $\xi \leftarrow \tau \xi + (1-\tau) \theta$.

Let $I \in \mathbb{R}^{W \times H \times C}$, $\mathbb{T}_1$ and $\mathbb{T}_2$ denote a training image and two sets of image augmentation strategies, respectively. 
Our method generates two augmented views $v_1 = t_1 (I)$ and $v_2 = t_2 (I)$ from $I$ by applying different image augmentations $t_1 \in \mathbb{T}_1$ and $t_2 \in \mathbb{T}_2$.
These augmented images ($v_1, v_2$) are respectively fed into the two encoder networks ($f_{\theta}, f_{\xi}$) having the last Global Average Pooling (GAP) layer removed to obtain spatial feature maps $m_1 = f_{\theta}(v_1) \in \mathbb{R}^{\widetilde{W} \times \widetilde{H} \times \widetilde{C}}$ and the same size of $m_2 = f_{\xi}(v_2)$.

Unlike previous methods that minimize the global representation distance between aggressive augmented $v_1$ and $v_2$ regardless of semantic information, we propose a method to minimize the local representation distance between the two local regions only if they associated with the same spatial regions and thus the same semantic meanings.

To do so, as shown in Figure \ref{fig:scrl_architecture}, we first find the intersection regions $ \mathcal{IS}(v_1, v_2) \in \mathbb{R}^{ \widehat{W} \times \widehat{H} \times C} $ on $I$, where $\mathcal{IS}(\cdot)$ denotes an operation that generates a spatially corresponding region between $v_1$ and $v_2$, and $\widehat{W}$ and $\widehat{H}$ are width and height of it, respectively.

After finding the intersection region, we randomly sample an arbitrary box $B=(x, y, w, h)$ in $\mathcal{IS}$ such that
\begin{equation}
\begin{split}
    \label{eqn:box_gen}
    w \sim \text{Unif}\: (\widehat{W}_{\textit{min}}, \widehat{W}),\;\; &
    x \sim \text{Unif}\: (0, \widehat{W}-w),\\
    h \sim \text{Unif}\: (\widehat{H}_{\textit{min}}, \widehat{H}),\;\; &
    y \sim \text{Unif}\: (0, \widehat{H}-h),
\end{split}
\end{equation}
where $\widehat{W}_{\textit{min}}=W/\widetilde{W}$ and $\widehat{H}_{\textit{min}}=H/\widetilde{H}$ (\eg, we use $\widehat{W}_{\textit{min}}=\widehat{H}_{\textit{min}}=32$ for ImageNet training with ResNet-50 and ResNet-101).
Then, to be utilized for spatial representation matching, $B$ has to be translated to the coordinates in each view $v_{i\in \{1,2\}}$, which can be denoted as $B_i=(x_i, y_i, w_i, h_i)$.

Due to aggressive image augmentations, the size, location, and internal color of each box ($B_1, B_2$) may be different for each views ($v_1, v_2$), however the semantic meaning in the cropped box area does not change between $B_1$ and $B_2$. Here, we crop a local region by a rectangular box. Therefore, in order to exactly map one rectangular box to another rectangular box after geometrical transformations, we exclude certain affine transformations such as shear operations and rotations.

Even without internal color changes, in general, the generated spatial feature maps from conventional CNNs are not coherently changed in scale and internal object translation of an input image \cite{SNIP}. We observe that the previous self-supervised learning methods based on the global consistency loss also have the same limitation in the spatial consistency.

Subsequently, to obtain the equally-sized local representations from $B_1$ and $B_2$, we crop the corresponding sample regions, called region-of-interests (RoIs), not on the input images but on the spatial feature maps and locally pool the cropped feature maps by 1x1 RoIAlign \cite{MaskRCNN}.

If the number of the boxes that are wanted to be sampled is more than one, we can efficiently obtain multiple pairs of local representations simultaneously by this cropping and pooling on a given spatial feature map, \ie, $p_i^{k} = \text{RoIAlign}(B_i^{k}, m_i)$, where $k=\{1,..,K\}$ and $K$ is the total number of generated boxes in an image.  

Within the online network, we then perform the projection, $z_1^{k}=g_{\theta}(p_1^{k})$, from the pooled representation $p_{1}^{k}$, followed by the prediction, $q_{\theta} (z_1^{k})$. At the same time, the target network outputs the target projection from $p_2^{k}$ such that $z_2^{k} = g_{\xi} (p_2^{k})$. Our spatial consistency loss is then defined as a mean squared error between the normalized prediction and the normalized target projection as follows

\vspace{-5mm}
\begin{equation}
\begin{split}
    \label{eqn:scrl_loss}
    \mathcal{L}_{\theta}^{\textit{SCRL}} = \frac{1}{K} { \sum_{k=1}^K {\lVert} \overline{q_{\theta} (z_1^k )} - \overline{z_2^k} {\rVert}_2^2 },
\end{split}
\end{equation}
where $\overline{q_{\theta} (z_1^k )} = q_{\theta} (z_1^k ) / {\lVert}q_{\theta} (z_1^k ){\rVert}_2$ and $\overline{z_2^k} = {z_2^k} / {\lVert} {z_2^k} {\rVert}_2$.
To symmetrize the loss $\mathcal{L}_{\theta}^{\textit{SCRL}}$ in Eq. \ref{eqn:scrl_loss}, we also feed $v_2$ to the online network and $v_1$ to the target network respectively to compute $\widetilde{\mathcal{L}}_{\theta}^{\textit{SCRL}}$ and the total loss is defined as

\vspace{-5mm}
\begin{equation}
\begin{split}
    \label{eqn:total_loss}
    \mathcal{L}_{\theta} = \mathcal{L}_{\theta}^{\textit{SCRL}} + \widetilde{\mathcal{L}}_{\theta}^{\textit{SCRL}}.
\end{split}
\end{equation}
During the self-supervised learning, we only optimize the online network to minimize $\mathcal{L}_{\theta}$ with respect to $\theta$. It is noted that we follow the BYOL framework for its simplicity in the use of only positive pairs without mode collapse. However, our spatial consistency strategy can be combined with general contrastive learning that also makes use of the negative pairs. We leave the explicit use of negative pairs for future works. 

There are lots of possible positive RoI pairs in a single image, and more diversely generated boxes can lead to efficient training as well as performance improvement. Thus, we promote the diversity of box instances by taking overlapped area among them into consideration. In detail, we compute the IoU (Intersection-over-Union) among the sampled RoIs and reject a candidate box if the IoU with previously generated boxes is larger than 50\%. We repeat this until the number of survived samples reaches to $K$. By default, we set $K=10$. The performance variations according to $K$ and whether the use of IoU thresholding will be presented in Section \ref{sec:exp}.

\subsection{Implementation Details}
\label{sec:impl_detail}

{\bf \noindent Dataset} For the task of self-supervised pretraining, we make use of 1.2 million training images on ImageNet \cite{ImageNet} as unlabeled data.

{\bf \noindent Image Augmentations} SCRL uses the same set of image augmentations in SimCLR \cite{SimCLRv1} and BYOL \cite{BYOL}. 
We perform simple random cropping \footnote{A random patch of the image is selected, with an area uniformly sampled between 20\% and 100\% of that of the original image, and an aspect ratio logarithmically sampled between 3/4 and 4/3.} with 224$\times$224 resizing, horizontal random flip, followed by a color distortion, and an optional grayscale conversion. 
Then, Gaussian blur and solarization are applied randomly to the images.

{\bf \noindent Network Architecture} We use a residual network \cite{ResNet} with 50 layers as our base networks $f_{\theta}$ and $f_{\xi}$. 
We also use ResNet-101 as a deeper network. 
Specifically, the feature map $m_i$ corresponds to the output of the last convolution block in ResNet, which has a feature dimension of (7, 7, 2048). 
After 1x1 RoIAlign \cite{MaskRCNN} of the randomly generated 10 RoIs, a feature dimension of 2048 is fed into projection $g_{\theta}$ that consists of a linear layer with output size 4096 followed by batch normalization, rectified linear units (ReLU) \cite{ReLU}, and a final layer with output dimension 256 as in BYOL \cite{BYOL}. 
The architecture of the predictor $q_{\theta}$ is the same as $g_{\theta}$.

{\bf \noindent Optimization} We use the same optimization method as in SimCLR \cite{SimCLRv1} and BYOL \cite{BYOL} (LARS optimizer \cite{LARS} with a cosine learning rate decay \cite{CosineLR} over 1000 epochs, warm-up period of 10 epochs, linearly scaled initial learning rate \cite{LinearLR}). 
We use the initial learning rate as 0.45.
The exponential moving average parameter $\tau$ is initialized as 0.97 and is increased to one during training. 
We use a batch size of 8192 on 32 V100 GPUs.

\begin{table}
\small
\centering
\begin{tabular}{c|ccc}
pretrain & AP & $\text{AP}_{50}$ & $\text{AP}_{75}$ \\ \hline \hline
random & 32.0 & 56.7 & 31.3 \\
supervised-IN & 53.2 & 81.7 & 58.2 \\
BYOL & 55.0 & 83.1 & 61.1 \\ \hline
\textbf{SCRL} & \textbf{57.2} & \textbf{83.8} & \textbf{63.9} \\ \hline
\end{tabular}
\vspace{1mm}
\caption{VOC detection using Faster R-CNN w/ FPN, ResNet-50. Supervised-IN denotes the representation trained using image labels on ImageNet (IN) dataset.}
\label{tab:voc}
\end{table}

\section{Experiments}
\label{sec:exp}

In this section, we elaborate the experiments on transferability to the localization tasks with various pre-trained models including ours. 
In addition, we conduct extensive ablation studies to understand the key factors in the proposed algorithm.

\subsection{Transfer to Localization Vision Tasks}

A main goal of representation learning is to learn features that are transferrable to downstream tasks. 
In this section, we compare SCRL with ImageNet supervised pretraining as well as the state-of-the-art self-supervised learning methods using ImageNet dataset without labels, transferred to various downstream localization tasks on PASCAL VOC \cite{VOC}, COCO \cite{COCO}, and CityScapes \cite{CityScapes}. 

As in MoCo \cite{MoCov1}, we fine-tune with synchronized BN \cite{SyncBN} that is trained, instead of freezing it \cite{ResNet}. We also use SyncBN in the newly initialized layers (\eg, FPN \cite{FPN}). 
Weight normalization is performed when fine-tuning supervised as well as unsupervised pretraining models. 
We use a batch size of 16 on 8 V100 GPUs.
Unless otherwise noted, all of the following experiments use the default hyper-parameters introduced in \texttt{Detectron2} \cite{detectron2}, which are more favorable hyper-parameters for the ImageNet supervised pretraining. 
Nonetheless, SCRL shows significant performance improvements over the ImageNet supervised pretraining as well as the state-of-the-art unsupervised learning methods.
Considering the performance gaps among the previous self-supervised learning methods including supervised image-level pretraining, the obtained performance improvements by SCRL are relatively large and significant across various tasks and networks.

\subsubsection {PASCAL VOC Object Detection}
\label{sec:voc_od}

We first evaluate SCRL on PASCAL VOC \cite{VOC} object detection task with ResNet-50-FPN \cite{ResNet, FPN} and Faster R-CNN \cite{FasterRCNN}.
We fine-tune all layers end-to-end with SyncBN \cite{SyncBN} and an input image is in [480, 800] pixels during training and 800 at inference.
The reported results in Table \ref{tab:voc} use the same experimental setting.
We use VOC \texttt{trainval07+12} as a training set and evaluate on VOC \texttt{test2007} set.
We evaluate more rigorous metrics of COCO-style AP and $\text{AP}_{75}$ including the default VOC metric of $\text{AP}_{50}$.

As shown in Table \ref{tab:voc}, SCRL substantially outperforms on both supervised pretraining and BYOL, \eg, by 4.0 points and 2.2 points in AP respectively.
Moreover, considering that the performance gap is more increased in terms of $\text{AP}_{75}$, SCRL contributes more to the correct box regression than the correct object classification.
We conjecture that spatially consistent matching in the upstream task improves its localization performance without any auxiliary technique and results in better box regression.

\begin{table}
\small
\centering
\begin{tabular}{c|c|ccc}
downstream & pretrain& AP & $\text{AP}_{50}$ & $\text{AP}_{75}$ \\ \hline \hline

\multirow{7}*{\shortstack{FRCNN \\ w/ FPN}} & random & 29.8 & 48.3 & 31.8 \\
& supervised-IN & 38.5 & 59.8 & 41.5 \\
\cline{2-5} & MoCo v2\color{red}$^{\dagger}$ & 37.1 & 57.2 & 40.2 \\
& SimCLR v2\color{red}$^{\dagger}$ & 38.1 & 58.9 & 41.3 \\
& SwAV\color{red}$^{\dagger}$ & 39.6 & 61.3 & 43.2 \\
& BYOL & 40.0 & 61.3 & 43.6 \\
\cline{2-5} & \textbf{SCRL} & \textbf{40.9} & \textbf{62.5} & \textbf{44.5} \\ \hline

\multirow{7}*{\shortstack{RetinaNet \\ w/ FPN}}
& random & 24.5 & 39.1 & 26.0 \\
& supervised-IN & 37.5 & 57.0 & 39.9 \\
\cline{2-5} & MoCo v2\color{red}$^{\dagger}$ & 37.0 & 55.8 & 39.5 \\
& SimCLR v2\color{red}$^{\dagger}$ & 37.4 & 56.5 & 40.3 \\
& SwAV\color{red}$^{\dagger}$ & 36.7 & 56.3 & 39.3 \\
& BYOL & 37.7 & 57.7 & 40.3 \\
\cline{2-5} & \textbf{SCRL} & \textbf{39.0} & \textbf{58.7} & \textbf{41.9} \\ \hline

\end{tabular}
\vspace{1mm}
\caption{COCO detection using ResNet-50. $^{\dagger}$: We use publicly available checkpoints released by the paper authors. For the upstream self-supervised pretraining task on ImageNet, MoCo v2 and SwAV are trained with \textit{800 epochs} while we run \textit{1000 epochs} for SimCLR v2, BYOL, and SCRL.}
\vspace{-1mm}
\label{tab:coco_res50}
\end{table}

\subsubsection {COCO Object Detection}
\label{sec:coco_od}

We evaluate SCRL on COCO object detection task. 
Faster R-CNN \cite{FasterRCNN} and RetinaNet \cite{FocalLoss} with FPN \cite{FPN} are used to measure the transferability within both two-stage and single-stage object detection frameworks. An input image is in [640, 800] pixels during training and is 800 at inference. We fine-tune all layers end-to-end and train on \texttt{train2017} set and evaluate on \texttt{val2017} set. 

As shown in Table \ref{tab:coco_res50}, SCRL outperforms with a significant margin not only the supervised pretraining on ImageNet but also the state-of-the-art self-supervised counterparts including MoCo v2 \cite{MoCov2}, SimCLR v2 \cite{SimCLRv2}, SwAV \cite{SwAV}, and BYOL \cite{BYOL}, in all metrics with both two-stage and single-stage detection architectures.

\begin{table}
\small
\centering
\begin{tabular}{c|ccc}
pretrain & AP & $\text{AP}_{50}$ & $\text{AP}_{75}$ \\ \hline \hline
random & 31.2 & 49.6 & 33.7 \\
supervised-IN & 40.4 & 61.6 & 44.1 \\ 
BYOL & 41.1 & 62.3 & 45.0 \\ \hline
\textbf{SCRL} & \textbf{42.9} & \textbf{64.4} & \textbf{46.8} \\ \hline
\end{tabular}
\vspace{1mm}
\caption{COCO detection using Faster R-CNN, ResNet-101-FPN.}
\label{tab:frcnn_res101}
\end{table}

We also perform evaluation using ResNet-101-FPN with Faster R-CNN. The evaluation results on COCO \texttt{val2017} in Table \ref{tab:frcnn_res101} show the same trend even if the encoder network is changed. 
While the upstream tasks using ResNet-50 are trained with 1000 epochs, both BYOL and SCRL of ResNet-101 are trained with 200 epochs for upstream tasks to measure the relative performance gap.

\begin{table}
\small
\centering
\setlength{\tabcolsep}{0.46em}
\begin{tabular}{c|ccc|ccc}
pretrain & $\text{AP}^{mk}$ & $\text{AP}_{50}^{mk}$ & $\text{AP}_{75}^{mk}$ & $\text{AP}^{bb}$ & $\text{AP}_{50}^{bb}$ & $\text{AP}_{75}^{bb}$ \\ \hline \hline
random & 28.7 & 46.9 & 30.6 & 30.9 & 49.7 & 33.2 \\
supervised-IN & 35.4 & 56.7 & 38.1 & 39.0 & 59.9 & 42.9 \\ 
MoCo \cite{MoCov1} & 35.1 & 55.9 & 37.7 & 38.5 & 58.9 & 42.0 \\
VADeR \cite{VADeR} & 35.6 & 56.7 & 38.2 & 39.2 & 59.7 & 42.7 \\
BYOL & 37.2 & 58.8 & 39.8 & 40.4 & 61.6 & 44.1 \\ \hline
\textbf{SCRL} & \textbf{37.7} & \textbf{59.6} & \textbf{40.7} & \textbf{41.3} & \textbf{62.4} & \textbf{45.0} \\ \hline
\end{tabular}
\vspace{1mm}
\caption{COCO instance segmentation using Mask R-CNN w/ FPN, ResNet-50: bounding-box AP ($\text{AP}^{bb}$) and mask AP ($\text{AP}^{mk}$) evaluated on \texttt{val2017}.}
\label{tab:coco_seg}
\end{table}

\begin{table}
\small
\centering
\begin{tabular}{c|ccc|cc}
\multirow{2}*{pretrain} & \multicolumn{3}{c|}{\shortstack{COCO\\keypoints detection}} & \multicolumn{2}{c}{\shortstack{Cityscapes\\segmentation}} \\
\cline{2-6} & AP & $\text{AP}_{50}$ & $\text{AP}_{75}$ & AP & $\text{AP}_{50}$ \\ \hline \hline
random & 63.2 & 85.3 & 68.9 & 26.2 & 52.1 \\
supervised-IN & 65.7 & 87.1 & 71.7 & 32.7 & 60.1 \\
BYOL & 65.8 & 87.0 & 72.0 & 34.2 & 62.1 \\ \hline
\textbf{SCRL} & \textbf{66.5} & \textbf{87.8} & \textbf{72.3} & \textbf{34.7} & \textbf{63.6} \\ \hline
\end{tabular}
\vspace{1mm}
\caption{COCO keypoints detection and Cityscapes instance segmentation using Mask R-CNN w/ FPN, ResNet-50}
\label{tab:coco_key_cityscapes}
\end{table}

\subsubsection{Other Localization Downstream Tasks}

We perform other localization downstream tasks with the default hyper-parameters as in \cite{detectron2} unless otherwise specified.

{\bf \noindent COCO Instance Segmentation. }
We fine-tune ResNet-50 model with Mask R-CNN \cite{MaskRCNN}. All experimental settings for Mask R-CNN are same as Section \ref{sec:coco_od}.
Similar to COCO detection, SCRL performs the best in all metrics not only $\text{AP}^{mk}$ but also $\text{AP}^{bb}$ in Mask R-CNN, as shown in Table \ref{tab:coco_seg}.

{\bf \noindent COCO Keypoints Detection. }
We use Mask R-CNN (keypoints version) with ResNet-50-FPN implemented in \cite{detectron2}, fine-tuned on COCO \texttt{train2017} and evaluated on \texttt{val2017}.
Table \ref{tab:coco_key_cityscapes} shows that SCRL outperforms in all metrics over the supervised pretraining as well as BYOL.

{\bf \noindent Cityscapes Instance Segmentation. }
On Cityscapes instance segmentation task \cite{CityScapes}, we fine-tune a model with Mask R-CNN. An input image is in [800, 1024] pixels during training and 1024 at inference. 
Unlike above downstream tasks, we use a batch of 8 which is the default training setting in \cite{detectron2}.
As shown in Table \ref{tab:coco_key_cityscapes}, SCRL outperforms the supervised ImageNet pretraining and BYOL.

\begin{table}
\small
\centering
\begin{tabular}{c|c|ccc}
pretrain & \# of RoIs & AP & $\text{AP}_{50}$ & $\text{AP}_{75}$ \\ \hline \hline
BYOL & N/A & 38.3 & 59.7 & 41.2 \\ \hline
SCRL & 1 & 39.2 & 60.3 & 42.4 \\
SCRL & 5 & 39.8 & 61.2 & 43.4 \\
SCRL & 10 & 40.2 & 61.3 & 43.7 \\ \hline
\end{tabular}
\vspace{1mm}
\caption{Performances according to the number of boxes used for spatial matching{\color{red}$^{\ddagger}$}.}
\label{tab:ablation_rois}
\end{table}

\begin{table}
\small
\centering
\begin{tabular}{c|c|c|ccc}
\# of RoIs & box jitter & IoU thr. & AP & $\text{AP}_{50}$ & $\text{AP}_{75}$ \\ \hline \hline
1 & none & 0.5 & 39.2 & 60.3 & 42.4 \\
1 & 10\% & 0.5 & 39.1 & 60.1 & 42.5 \\
1 & 20\% & 0.5 & 38.8 & 60.0 & 42.2 \\ \hline
10 & none & 0.5 & 40.2 & 61.3 & 43.7 \\
10 & 10\% & 0.5 & 40.0 & 61.0 & 43.5 \\
10 & 20\% & 0.5 & 39.6 & 60.8 & 43.1 \\
10 & $\infty$ & 0.5 & 38.8 & 59.8 & 42.2 \\ \hline
10 & none & none & 39.8 & 61.0 & 43.5 \\ \hline
\end{tabular}
\vspace{1mm}
\caption{Performances according to different box generations in SCRL{\color{red}$^{\ddagger}$}.}
\vspace{-1mm}
\begin{flushleft}
\footnotesize{$^{\ddagger}$: We run all experiments with \textit{200 epochs} for upstream task and fine-tune on COCO detection with the default parameters in \cite{detectron2}}
\end{flushleft}
\label{tab:ablation_spatial_match}
\end{table}

\vspace{3mm}
\subsection{Ablation Studies}

In this subsection, we delve deeper into our SCRL by performing various ablation studies. We run all ablations using 1000 epochs during upstream pretraining if not specified, and fine-tuned on COCO detection with Faster RCNN and ResNet-50-FPN.

\subsubsection{Number of Boxes Used for Spatial Matching}
As shown in Table \ref{tab:ablation_rois}, increasing the number of RoI pairs to be matched between the two views improves the downstream performances. In particular, SCRL outperforms BYOL even when using a single pair for the spatial matching. This ensures a fair comparison between the two methods from the perspective of the number of matched pairs per an image. It demonstrates that our method brings more effect than that from simply increasing the number of augmented samples. Another benefit from our approach is that increasing the number of matched pairs from a single image only adds negligible computational costs since they share the same feature map, not requiring multiple forward feature extractions as in multi-crop \cite{SwAV}.

\subsubsection{Importance of SCRL's Box Generation}
To evaluate the benefit of the precise box matching scheme, we perform experiments on what happens when relaxing the exact matching rule with respect to the two cropped regions in a pair.
In other words, we randomly jitter the matched boxes to different degrees, where each box can shrink or expand its width and height, and move its position within the margin of a specific percentage.
As shown in Table \ref{tab:ablation_spatial_match}, the heavier jittering incurs larger performance degradation, which supports the importance of exact spatial coherence induced to the feature space in SCRL. This is especially compared with BYOL which generates a randomly cropped pair on the input space regardless of its semantic coherence. 
We also confirm the effect of the technique, IoU thresholding, among the sampled regions in that it prevents the creation of redundant boxes and produces more diverse non-overlapping boxes, which results in better spatial representations.

\subsubsection{Representation Power in Small Data Regime}

To confirm that the representation itself learned with SCRL contains a lot of more useful information for an object detection task, we evaluate the object detection performance fine-tuned with only 10\% of COCO \texttt{train2017} dataset while keeping the same hyper-parameters in \cite{detectron2}. As shown in Table \ref{tab:coco_fewshot}, SCRL outperforms all the baselines with noticeable margins. This indicates that SCRL provides a more transferable representation that can be harnessed for localization tasks even in a small data regime.

\begin{table}
\small
\centering
\begin{tabular}{c|ccc}
pretrain & AP & $\text{AP}_{50}$ & $\text{AP}_{75}$ \\ \hline \hline
random  & 17.8 & 32.0 & 17.9 \\
supervised-IN  & 22.6 & 38.4 & 23.5 \\ \hline
MoCo v2 & 20.9	& 34.8 & 21.7 \\
SimCLR v2 & 22.1 &	37.3 & 23.0 \\
SwAV & 25.5 & \textbf{43.3} & 26.4 \\
BYOL & 25.5 & 42.3 & 26.9 \\ \hline
\textbf{SCRL} & \textbf{26.4} & 43.2 & \textbf{28.0} \\ \hline

\end{tabular}
\vspace{1mm}
\caption{COCO detection with 10\% training dataset using Faster R-CNN w/ FPN, ResNet-50.}
\label{tab:coco_fewshot}
\end{table}

\subsubsection{Performance on Varied Downstream Schedules}

\begin{table}
\small
\centering
\begin{tabular}{c|ccccc}
\multirow{2}*{pretrain} & \multicolumn{5}{c}{LR schedule} \\
\cline{2-6} & 0.5$\times$ & 1$\times$ & 3$\times$ & 5$\times$ & 7$\times$ \\ \hline \hline
random & 23.2 & 29.8 & 38.5 & 40.3 & 40.9 \\
supervised-IN & 35.6 & 38.5 & 40.9 & 41.3 & 41.4 \\
BYOL & 37.0 & 40.0 & 41.6 & 41.7 & 41.6 \\ \hline
\textbf{SCRL} & \textbf{38.3} & \textbf{40.9} & \textbf{42.3} & \textbf{42.5} & \textbf{42.6} \\ \hline
\end{tabular}
\vspace{1mm}
\caption{Object detection $\text{AP}$ on COCO \texttt{val2017} with training schedules from 0.5$\times$ (45k iterations) to 7$\times$ (630k iterations). 
}
\label{tab:lr_schedule}
\end{table}

Table \ref{tab:lr_schedule} illustrates downstream performance under various lengths of the training schedule \ie, 0.5$\times$, 1$\times$, 3$\times$, 5$\times$, 7$\times$, in comparison to other baselines. As discussed in \cite{RethinkIN}, ImageNet pretraining shows rapid convergence than random initialization at the early stage of training  but the final performance is not any better than the model trained from scratch. Similarly, BYOL appears to provide a slightly better initial point but the gap between the aforementioned baselines wears off at last. 
On the other hand, SCRL goes beyond that limit and the noticeable gain is preserved even in longer schedules. 
We argue that SCRL provides task-specific representations of quality that the previous pretraining methods have not yet achieved. It also implies that one should rethink, in fact, not the ImageNet pretraining itself but the right way of doing it to the specific downstream task.

\begin{table}
\small
\centering
\setlength{\tabcolsep}{0.4em}
\begin{tabular}{c|c|cc}
pretrain & \footnotesize{\shortstack{upstream \\ epochs}} & \footnotesize{\shortstack{global linear eval. \\ (ImageNet)}} & \footnotesize{\shortstack{RoI linear eval. \\ (COCO GT-boxes)}} \\ \hline \hline

supervised-IN & 90 & 74.3 & 72.7\\ \hline
MoCo v2 & 800 & 71.1 & 69.3 \\
SimCLR v2 & 1000 & 71.7 & 71.7 \\
SwAV & 800 & \textbf{75.3} & 72.6 \\
BYOL & 1000 &  74.3 & 71.5  \\ \hline 
\textbf{SCRL} & 1000 & 70.3 & \textbf{74.8}\\ \hline
\end{tabular}
\vspace{1mm}
\caption{Linear evaluation accuracy on ImageNet and COCO GT boxes. Note that, for RoI linear evaluation, we start all over tracking the running statistics of the batch normalization layers during the linear head training to adapt to the unseen distribution of COCO dataset.}
\vspace{-1mm}
\label{tab:linear_eval}
\end{table}

\begin{figure*}
    \centering
    \includegraphics[height=.15\textwidth]{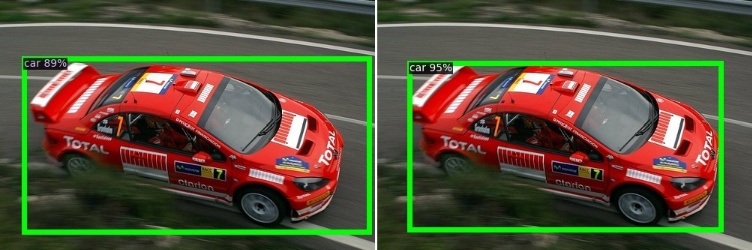}\hfill
    \includegraphics[height=.15\textwidth]{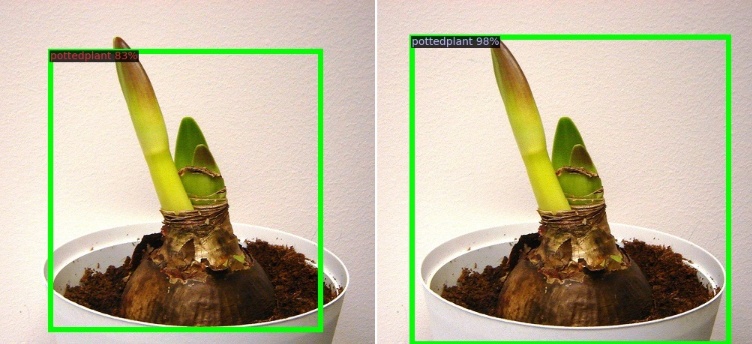}\hfill
    \includegraphics[height=.15\textwidth]{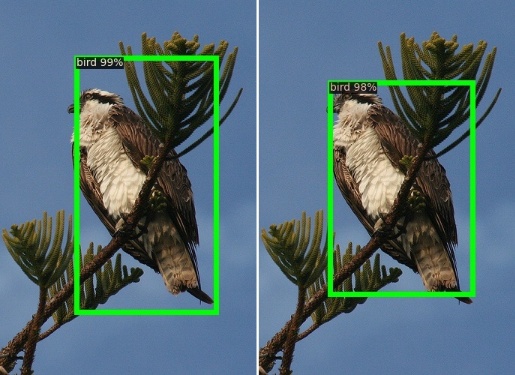}
    \\[\smallskipamount]

    \includegraphics[height=.15\textwidth]{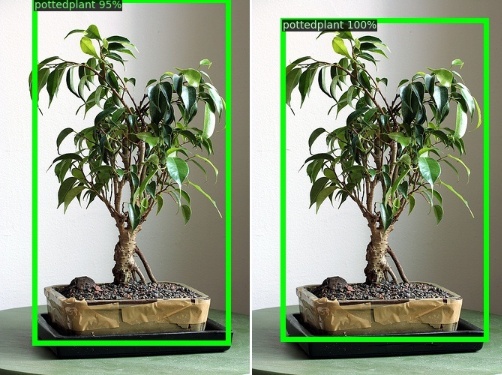}\hfill    \includegraphics[height=.15\textwidth]{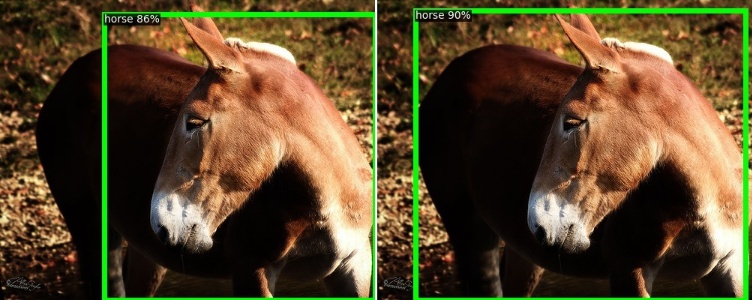}\hfill 
    \includegraphics[height=.15\textwidth]{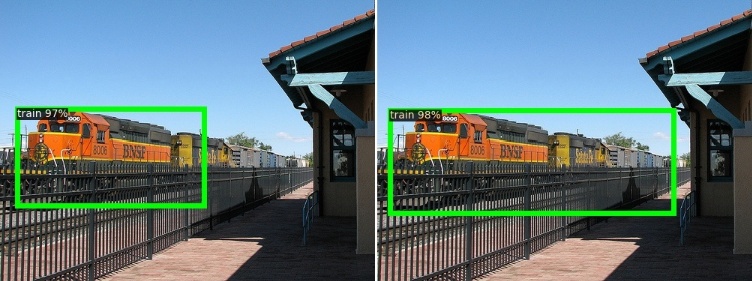}\hfill
    \\[\smallskipamount]    

    \includegraphics[height=.136\textwidth]{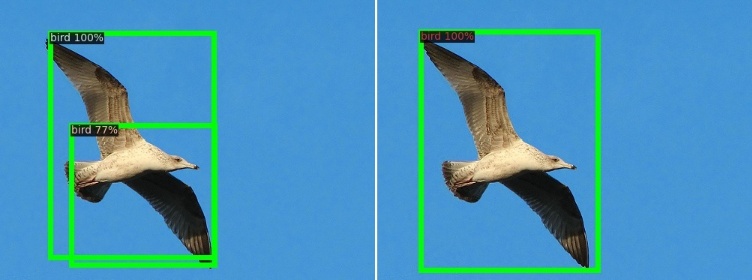}\hfill
    \includegraphics[height=.136\textwidth]{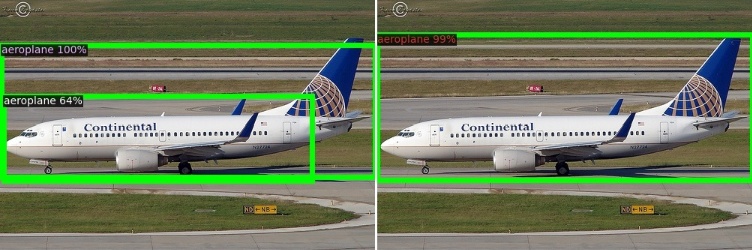}\hfill
    \includegraphics[height=.136\textwidth]{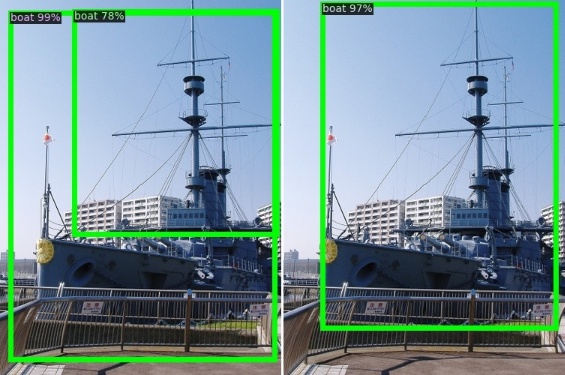}\hfill
    \\[\smallskipamount]
    
    \includegraphics[height=.136\textwidth]{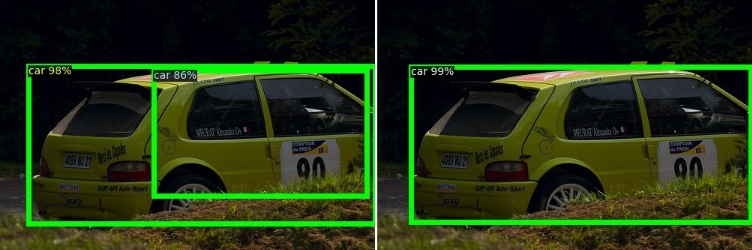}\hfill
    \includegraphics[height=.136\textwidth]{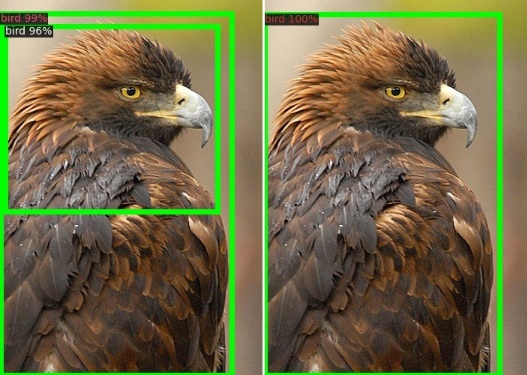}\hfill
    \includegraphics[height=.136\textwidth]{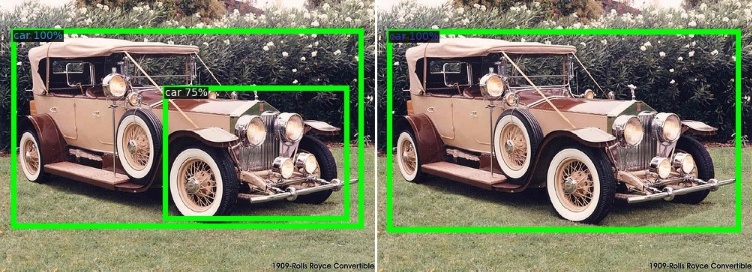}\hfill

    \caption{Qualitative comparison between BYOL (left) and SCRL (right) on PASCAL VOC detection w/ Faster R-CNN, ResNet-50-FPN. The first two rows show the regression capability of bounding boxes. The latter two rows present the malfunction of BYOL.} 
    \label{fig:qualitative}
\end{figure*}

\subsubsection{RoI Linear Evaluation using GT Boxes}

In this section, we propose a simple protocol for localization tasks, similar to the linear classification evaluation, but focusing only on the region-of-interests given by the ground-truth boxes of localization datasets. Specifically, we train a single linear layer, on the top of the frozen backbone followed by a RoIAlign layer delivering the localized features of ground-truth boxes to the learner\footnote{Considering the small-sized object, 800$\times$800 resized-images are used to train a single linear layer. We only perform horizontal random flip and use a batch size of 512 on 8 V100 GPUs. Initial LR is 0.1 with a cosine learning rate decay over 80 epochs, warm-up period of 5 epochs.}.

Table \ref{tab:linear_eval} shows linear evaluation results, using the pre-GAP features of ResNet-50, under the both standard and the proposed protocol. SCRL outperforms other baselines on our protocol. Interestingly, trade-off is observed between the two protocols when the performance is saturated, which suggests that, for object detection tasks, learned representations have to be linearly separable under spatial constraints rather than in a global fashion.

\vspace{3mm}
\subsection{Qualitative Analysis}
We found that the performance gain of our method comes more from the box regression than the category classification as we described in \ref{sec:voc_od}.
To understand exactly what aspect of the regression is improved, we scrutinize the detection result in qualitative perspective. 
Figure \ref{fig:qualitative} illustrates the detected boxes with correct class prediction, where the left and right figure of each pair represent the outcomes from the model having been initialized with BYOL and SCRL, respectively. 

The first two rows are the cases in which SCRL predicts precise and tight box boundaries while BYOL fails to do so. 
The latter two rows represent the ones where BYOL detects a smaller box embracing only a mere \textit{part} of the entire object and confidently predicts it as a whole whereas SCRL successfully resists it.

We conjecture that this misbehavior of BYOL pretraining attributes to its translation-invariant representations by matching features between two randomly cropped views, while SCRL does learn the features sensitive to positional variation. Both methods learn scale-invariant features but significantly different in that BYOL has a limited vision of already-cropped and resized-to-the-same-scale patches while SCRL sees the patches on the feature map along with the global context thanks to the peripheral vision from receptive field. Thereby, it may help SCRL to capture a full view of objects considering entire context regardless of its size and position.

\vspace{3mm}
\section{Conclusion}

In this paper, we have presented a novel self-supervised learning algorithm, SCRL, that tries to produce consistent spatial representations by minimizing the distance between spatially matched regions. 
By doing so, the proposed SCRL outperforms, with a significant margin, the state-of-the-art self-supervised learning methods on various downstream localization tasks.

{\small
\bibliographystyle{ieee_fullname}
\bibliography{egbib}
}


\clearpage

\section*{Appendix}
\addcontentsline{toc}{section}{Appendices}
\renewcommand{\thesubsection}{\Alph{subsection}}

\setcounter{table}{0}
\renewcommand{\thetable}{A\arabic{table}}

\setcounter{figure}{0}
\renewcommand{\thefigure}{A\arabic{figure}}

\subsection{Implementation details}

\subsubsection{Image Augmentations for SCRL}
We use the same set of image augmentations in SimCLR \cite{SimCLRv1} and BYOL \cite{BYOL} except random cropping. 
We crop the patch of the image with an area uniformly sampled between 20\% and 100\% of that of the original image as described in Section \ref{sec:impl_detail}. We observe that this change is not detrimental to BYOL and results in increasing the intersection area between SCRL's $v_1$ and $v_2$.
Table \ref{tab:image_aug} shows the image augmentation parameters from BYOL \cite{BYOL}.

\subsubsection{PASCAL VOC Object Detection}
We use Faster R-CNN \cite{FasterRCNN} with ResNet-50-FPN \cite{ResNet, FPN}. The base learning rate is set to 0.02 and multiplied by 0.1 at 12000 and 16000 steps of training, respectively. 
We train a model over 18000 steps with 16 batches.

\subsubsection{COCO Object Detection}
We use Faster R-CNN \cite{FasterRCNN} and RetinaNet \cite{FocalLoss} with ResNet-50-FPN \cite{ResNet, FPN}. The base learning rates are set to 0.02 for Faster R-CNN and 0.01 for RetinaNet, and multiplied by 0.1 at 60000 and 80000 steps of training, respectively. 
We train a model over 90000 steps with 16 batches.

In the case of training on various downstream schedules, we multiply the training schedule to the default setting, \ie milestone for 1/10 learning rate decaying step is [30000, 40000], and train a model over 45000 steps for $\times$0.5 LR schedule.

\subsubsection{COCO Instance Segmentation}
We use Mask R-CNN \cite{FasterRCNN} with ResNet-50-FPN \cite{ResNet, FPN}. The base learning rate is set to 0.02 and multiplied by 0.1 at 60000 and 80000 steps of training, respectively. 
We train a model over 90000 steps with 16 batches.

\subsubsection{COCO Keypoints Detection}
We use Mask R-CNN \cite{FasterRCNN} (keypoint version) with ResNet-50-FPN \cite{ResNet, FPN}. The base learning rate is set to 0.02 and multiplied by 0.1 at 60000 and 80000 steps of training, respectively. 
We train a model over 90000 steps with 16 batches.

\subsubsection{Cityscapes Instance Segmentation}
We use Mask R-CNN \cite{FasterRCNN} with ResNet-50-FPN \cite{ResNet, FPN}. The base learning rate is set to 0.01 and multiplied by 0.1 at 18000 steps of training.
We train a model over 24000 steps with 8 batches.

\begin{table}
\small
\centering
\begin{tabular}{ccc}
\hline
augmentation parameter & $\mathbb{T}_1$ & $\mathbb{T}_2$ \\ \hline \hline
random crop probability & 1.0 & 1.0 \\
flip probability & 0.5 & 0.5 \\
color jittering probability & 0.8 & 0.8 \\
brightness adjustment max intensity & 0.4 & 0.4 \\
contrast adjustment max intensity & 0.4 & 0.4 \\
saturation adjustment max intensity & 0.2 & 0.2 \\
hue adjustment max intensity & 0.1 & 0.1 \\
color dropping probability & 0.2 & 0.2 \\
Gaussian blurring probability & 1.0 & 0.1 \\
solarization probability & 0.0 & 0.2 \\ \hline
\end{tabular}
\vspace{1mm}
\caption{Parameters used to generate image augmentations \cite{BYOL}.}
\label{tab:image_aug}
\end{table}

\subsection{Representation Quality of FPN Evaluated under RoI Linear Protocol}
\label{sec:after-downstream}

We conduct the RoI linear evaluation with the ResNet-50-FPN backbone, that is pretrained on ImageNet, and verify the correlation between the RoI evaluation accuracy and the object detection AP after being fine-tuned on the downstream task. 
For downstream object detection, we use the Faster R-CNN method with ResNet-50-FPN on COCO dataset.
To make the representation for RoI evaluation compatible with FPN architecture, we concatenate the RoI-aligned features from every stage of the feature pyramid and feed it to a linear head as usual. Note that we use minibatch statistics for batch normalization layers during the training of the linear head in order to adapt to the statistics of COCO dataset, while simultaneously tracking running statistics that is to be used during the test. As shown in Table \ref{tab:after_downstream}, a strong positive correlation (Pearson's coefficient is 0.97) between the two columns is observed. This justifies our assumption on our protocol with which we can measure the quality of representation without direct access to object detection downstream task.

\begin{table}
    \centering
    \small
    \begin{tabular}{c|cc}
        upstream & \shortstack{AP on COCO \\ downstream} & \shortstack{RoI evaluation acc. \\ after downstream} \\ \hline \hline
        random & 29.77 & 68.52  \\
        supervised-IN & 38.52 & 79.20 \\ \hline
        MoCo-v2 & 37.12 & 77.04 \\
        SimCLR-v2 & 38.14 & 77.67 \\
        SeLa-v2 & 37.75 & 80.55 \\
        DeepCluster-v2 & 37.97 & 80.61 \\
        SwAV & 39.58 & 81.98 \\
        BYOL & 39.98 & 81.34 \\  \hline
        \textbf{SCRL} & \textbf{40.94} & \textbf{82.39} \\ \hline
    \end{tabular}\vspace{2mm}
    \caption{The correlation between RoI evaluation accuracy and AP after being fine-tuned on the downstream task.}
    \label{tab:after_downstream}
\end{table}

\begin{figure}
\small
    \centering
    \includegraphics[width=1.0\linewidth]{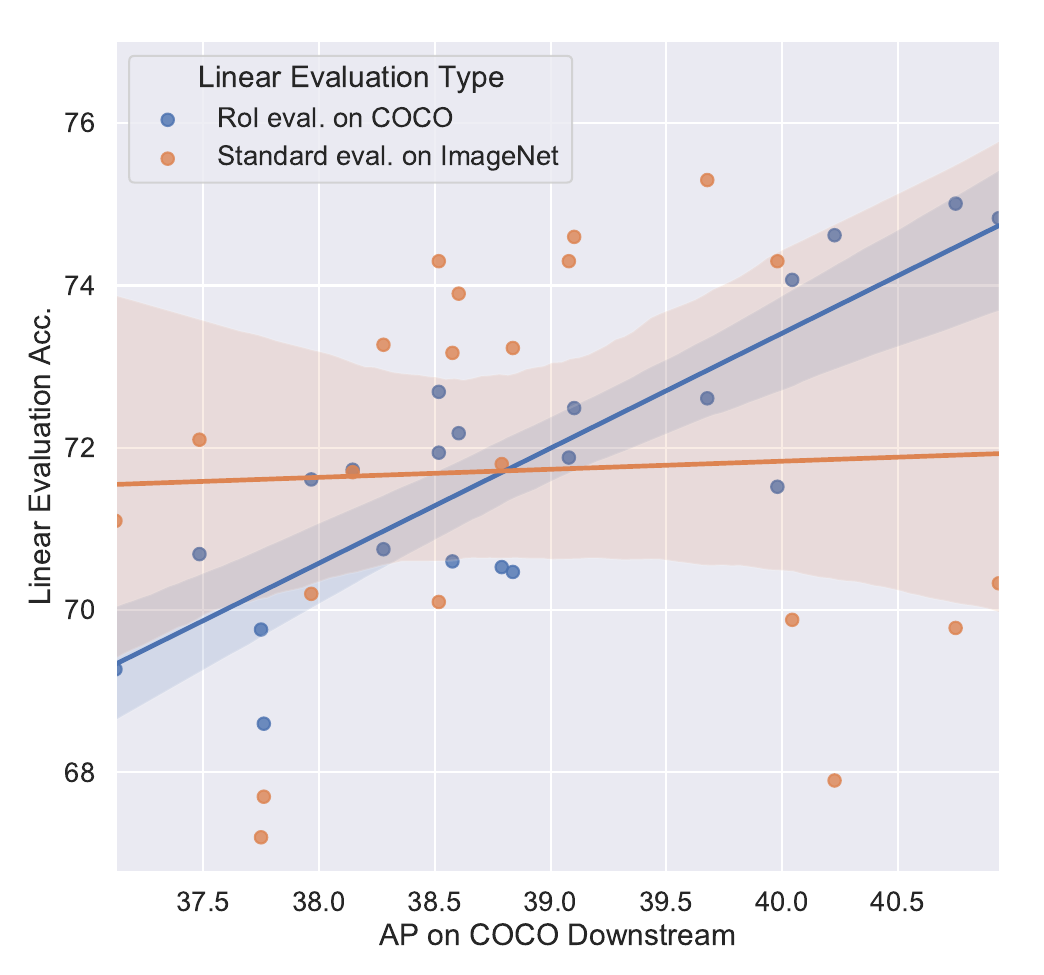}
    \caption{The correlation between two types of linear evaluation after upstream and the actual downstream performance using the initial representation. The proposed RoI evaluation (blue) shows higher positive correlation than the standard linear evaluation protocol (orange). Each point corresponds to different upstream methods with various upstream schedules. The straight line depicts linear regression result and the shaded area around it represents 95\% confidence interval. 
    }
    \label{fig:before_downstream}
\end{figure}

\subsection{The Correlation between Linear Evaluation Protocols and the Downstream Performance}

In this section, we further discuss how the proposed and the standard linear evaluation protocols are correlated to the actual downstream performance with a wider range of examples. Specifically, we use SeLa-v2 \cite{SwAV}, DeepCluster-v2 \cite{SwAV}, SimCLR-v2 \cite{SimCLRv2}, MoCo-v2 \cite{MoCov2}, SwAV \cite{SwAV}, BYOL \cite{BYOL}, and our method, SCRL, with varied upstream epochs and ablated optional techniques such as multicrop in SwAV or box generation details in SCRL. In the case of other baselines, publicly available checkpoints provided by the authors are used. For upstream, ResNet-50 backbone is pre-trained on ImageNet with different methods and, for downstream, Faster R-CNN with additional FPN is fine-tuned on COCO detection task. We apply the same treatment as described in Section \ref{sec:after-downstream} for batch normalization layers during RoI evaluation. 

As shown in Figure \ref{fig:before_downstream}, the proposed protocol, RoI evaluation shows a significantly higher correlation to the downstream performance and, in addition, Pearson correlation of our protocol (0.85) is 20$\times$ higher than the one of the standard image classification protocol on ImageNet (0.04). 
Based on this observation, we suggest that one can use our protocol to measure transferability to the object detection during upstream under self-supervision, in an online manner, without access to the actual downstream validation.

\subsection{Upstream training with COCO dataset}
We train the model on upstream task using unlabeled COCO \texttt{train2017} for the number of steps that correspond to 200 epochs on ImageNet.
Then, we fine-tune it on COCO detection task with 1$\times$ training schedule and obtain 39.0 AP, which is 3.6 points higher than BYOL as shown in Table {\ref{tab:coco_upstream}}.

\subsection{Downstream training with Sparse R-CNN}
We perform COCO detection task with Sparse R-CNN {\cite{SparseRCNN}} that does not use predefined anchors and non-maximum suppression (NMS) through the bipartite matching, similar to DETR {\cite{detr}}. 
As with other experiments we compared in the paper, SCRL outperforms the supervised ImageNet pre-trained counterpart on Sparse R-CNN.
This experiment shows that our SCRL can be applicable to any other detection frameworks to boost the performance without additional training cost and efforts.

\begin{table}
\small
\centering
\begin{tabular}{c|c|ccc}
upstream dataset & pretrain & AP & $\text{AP}_{50}$ & $\text{AP}_{75}$ \\ \hline \hline
\multirow{2}*{COCO}
 & BYOL & 35.4 & 55.6 & 38.0 \\ 
 & \textbf{SCRL} & \textbf{39.0} & \textbf{60.1} & \textbf{42.4} \\ \hline
\end{tabular}
\vspace{1mm}
\caption{COCO detection using Faster R-CNN, ResNet-50-FPN. Upstreams are trained with the unlabeled COCO dataset with 2000 epochs.}
\label{tab:coco_upstream}
\end{table}

\begin{table}
\small
\centering
\begin{tabular}{c|c|ccc}
method & pretrain & AP & $\text{AP}_{50}$ & $\text{AP}_{75}$ \\ \hline \hline
\multirow{2}*{\shortstack{Sparse R-CNN}}
 & \footnotesize{supervised-IN} & 42.3 & 61.2 & 45.7 \\ 
 & \textbf{SCRL} & \textbf{44.3} & \textbf{63.0} & \textbf{48.0} \\ \hline
\multirow{2}*{\shortstack{Sparse R-CNN}$^{\star}$ }
 & \footnotesize{supervised-IN} & 44.5 & 63.4 & 48.2 \\ 
 & \textbf{SCRL} & \textbf{46.7} & \textbf{65.7} & \textbf{51.1} \\ \hline
\end{tabular}
\vspace{1mm}
\caption{COCO detection using Sparse R-CNN {\cite{SparseRCNN}}, ResNet-50-FPN.
The training schedule is 36 epochs and all downstream tasks are trained with the default hyper-parameters as in {\cite{SparseRCNN}}.
Here $^{\star}$ indicates that the model is trained with 300 learnable proposal boxes and random crop training augmentation, similar to Deformable DETR {\cite{DeformableDETR}}.}
\label{tab:sparse_rcnn}
\end{table}

\begin{figure*}
    \centering

    \includegraphics[height=.123\textwidth]{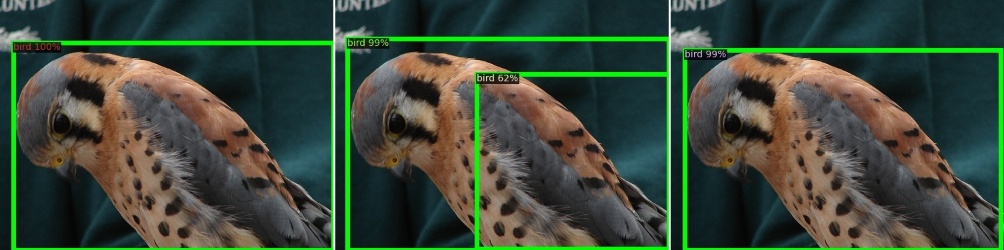}\hfill
    \includegraphics[height=.123\textwidth]{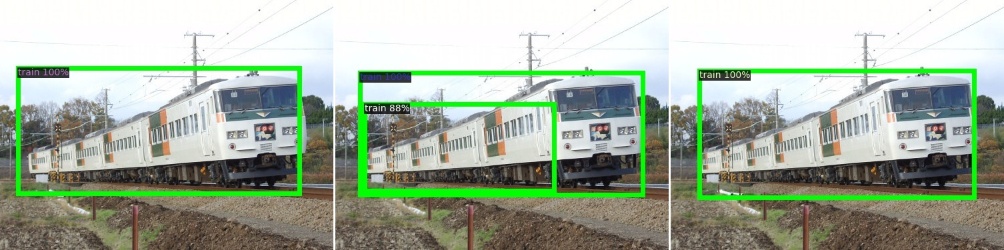}\hfill
    \\[\smallskipamount]

    \includegraphics[height=.146\textwidth]{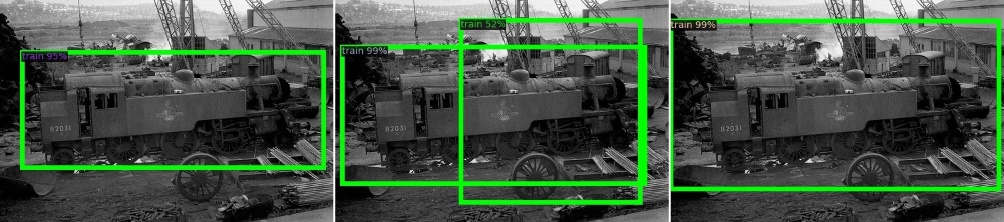}\hfill
    \includegraphics[height=.146\textwidth]{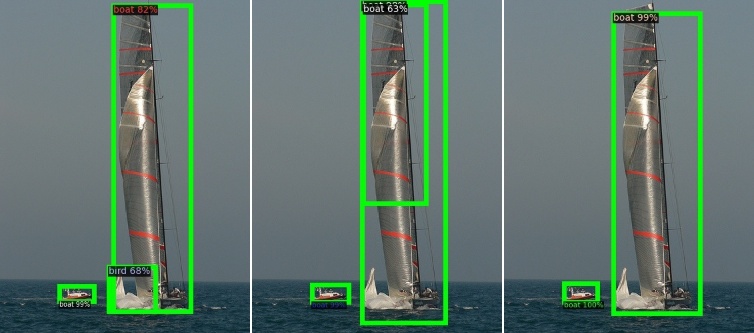}\hfill
    \\[\smallskipamount]

    \includegraphics[height=.153\textwidth]{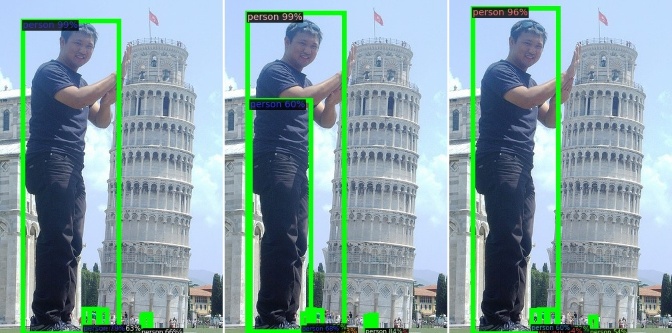}\hfill
    \includegraphics[height=.153\textwidth]{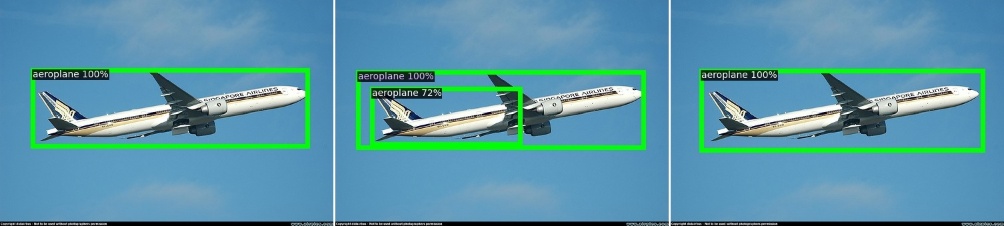}\hfill
    \\[\smallskipamount]

    \includegraphics[height=.146\textwidth]{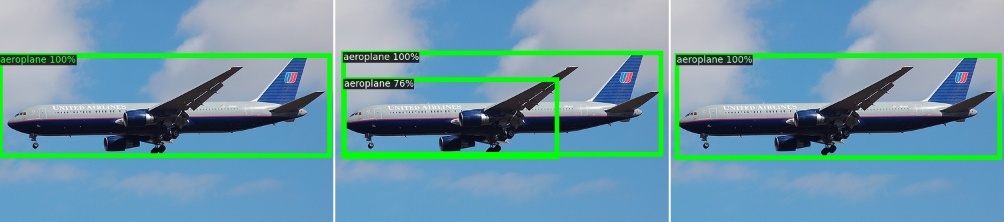}\hfill
    \includegraphics[height=.146\textwidth]{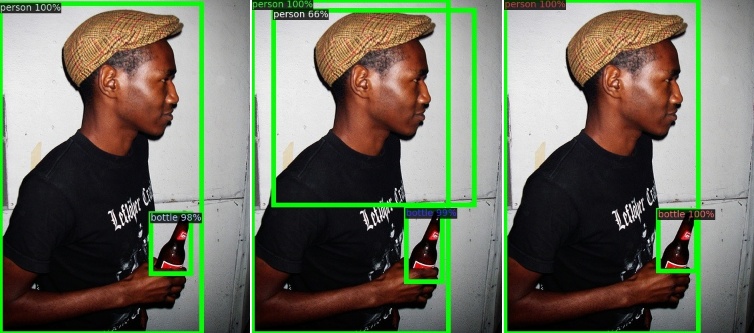}\hfill
    \\[\smallskipamount]

    \includegraphics[height=.113\textwidth]{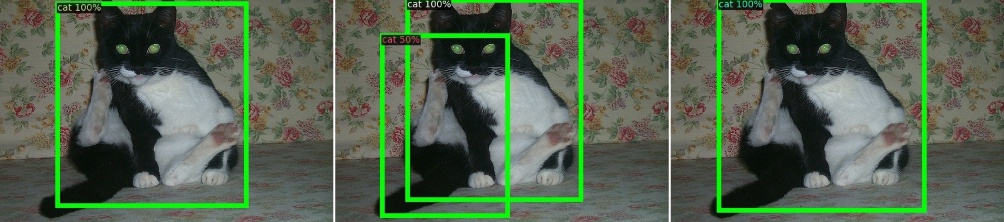}\hfill
    \includegraphics[height=.113\textwidth]{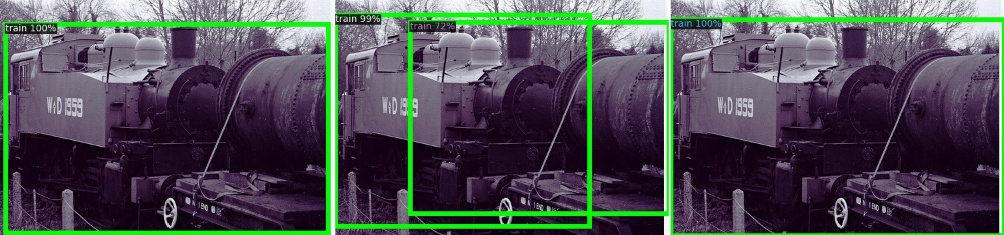}\hfill
    \\[\smallskipamount]

    \includegraphics[height=.151\textwidth]{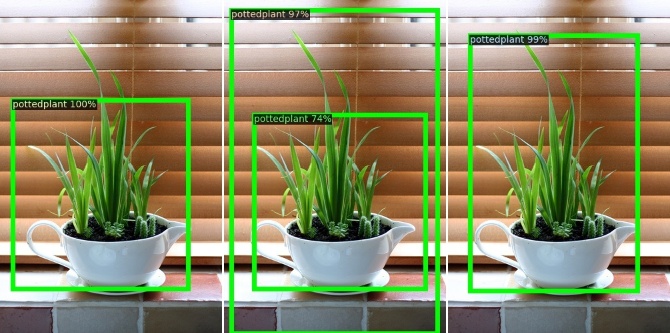}\hfill
    \includegraphics[height=.151\textwidth]{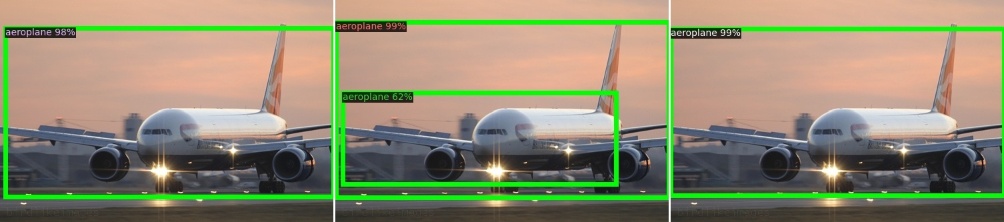}\hfill
    \\[\smallskipamount]

    \includegraphics[height=.164\textwidth]{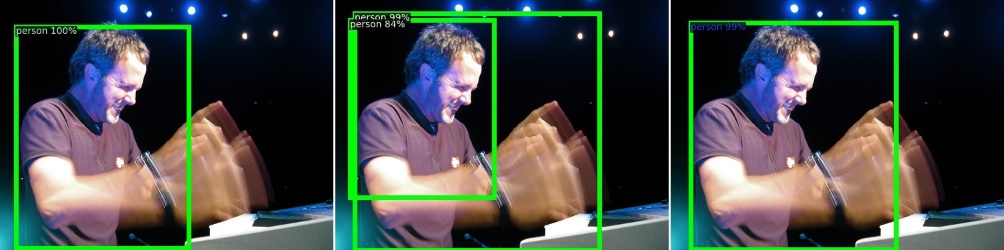}\hfill
    \includegraphics[height=.164\textwidth]{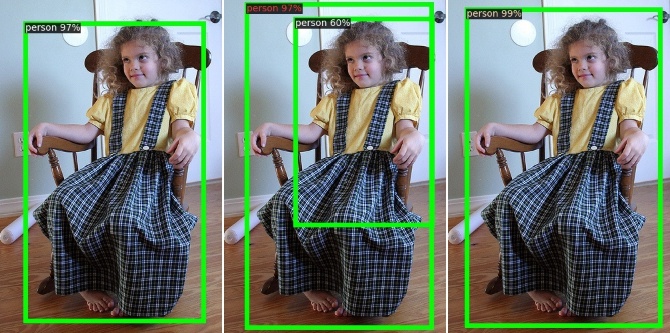}\hfill
    \\[\smallskipamount]

    \includegraphics[height=.149\textwidth]{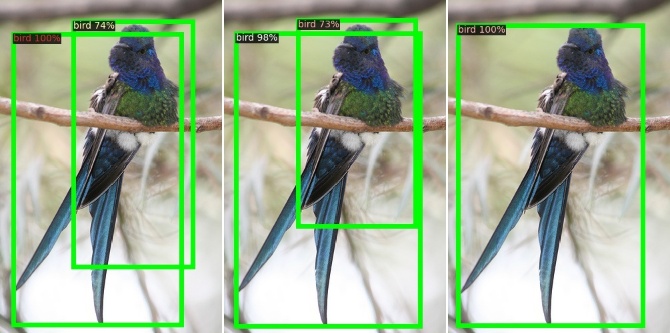}\hfill
    \includegraphics[height=.149\textwidth]{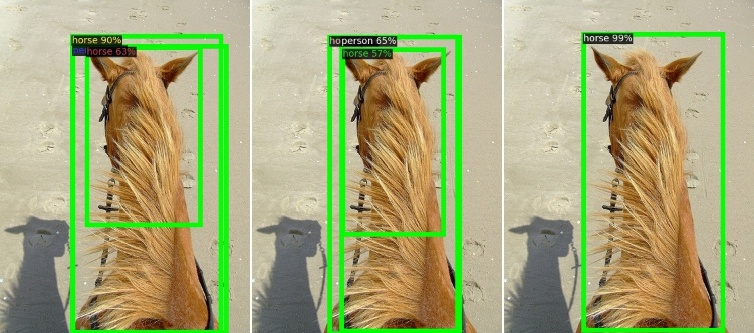}\hfill
    \includegraphics[height=.149\textwidth]{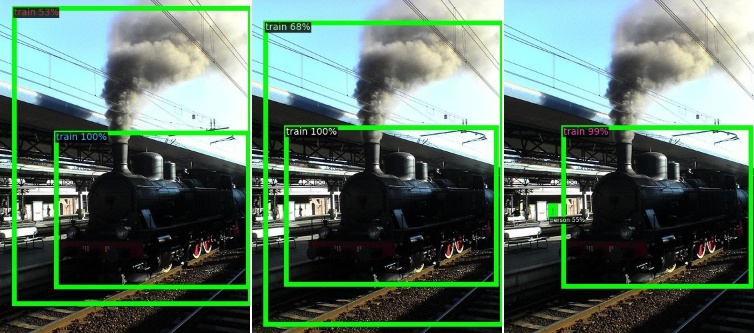}\hfill
    \\[\smallskipamount]
    
    \caption{Qualitative comparison among ImageNet (left), BYOL (middle) and SCRL (right) on PASCAL VOC detection w/ Faster R-CNN, ResNet-50-FPN.}
    \label{fig:qualitative_appendix}
\end{figure*}

\begin{figure*}
    \centering
    \includegraphics[height=.16\textwidth]{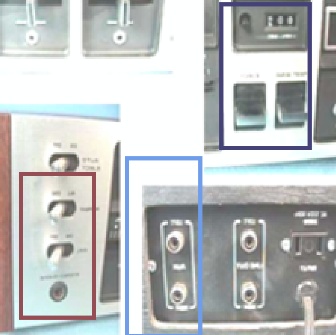}\hfill
    \hspace{-2mm}\includegraphics[height=.16\textwidth]{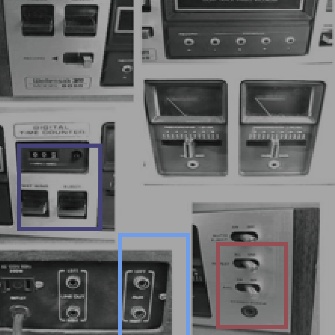}\hfill
    \includegraphics[height=.16\textwidth]{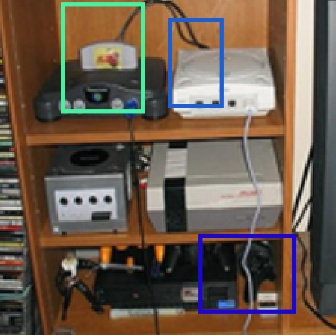}\hfill
    \hspace{-2mm}\includegraphics[height=.16\textwidth]{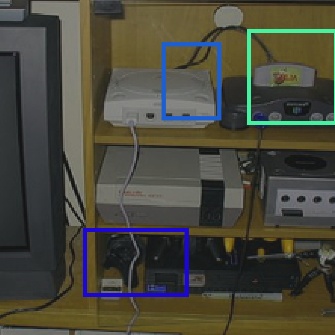}\hfill
    \includegraphics[height=.16\textwidth]{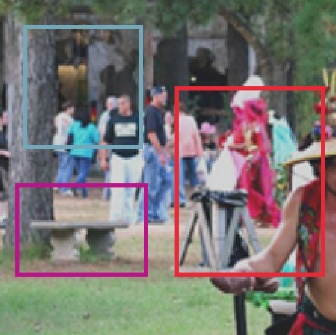}\hfill
    \hspace{-2mm}\includegraphics[height=.16\textwidth]{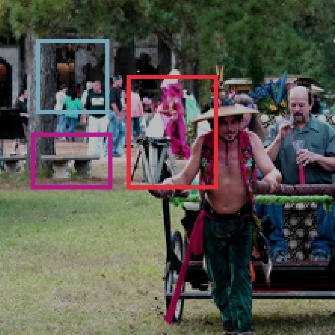}\hfill

    \caption{Randomly generated boxes from two augmented views. 
    In two augmented views, two rectangular regions of the same color are spatially matched.}
    \label{fig:random_boxes}
\end{figure*}

\subsection{Additional Qualitative Analysis}
Figure \ref{fig:qualitative_appendix} illustrates the detected boxes with correct class prediction, where the triplet-pair of each image represents the results from the model having been initialized with ImageNet pretraining, BYOL, and SCRL, respectively.
We found BYOL tends to detect a part of the object simultaneously as well as the entire object. 
As we described in the paper, we conjecture that these unintended consequences of BYOL are caused by semantically inconsistent matching between randomly cropped views by aggressive augmentation. 
Though BYOL outperforms ImageNet pre-trained representation on the entire test set, the shortcoming observed in this qualitative analysis implies that there still exists room for further improvement, which is exactly where SCRL tries to fill by introducing the spatial consistency.
Thereby, SCRL detects the entire object solidly since it produces position and scale-invariant features.
Interestingly, the bottom row in Figure \ref{fig:qualitative_appendix} shows that SCRL is robust to such phenomena even though when ImageNet pretraining generate multiple boxes in a single object.

\subsection{Random Boxes from Two Augmented Views}
Figure {\ref{fig:random_boxes}} shows randomly generated boxes from two augmented views during the upstream training.
We use $K=3$ which is the total number of generated boxes in an image for simplicity while the main experiment generates 10 boxes ({\ie} {\hspace{1mm}} $K=10$) as a default training setting.

\begin{table}
\small
\centering
\begin{tabular}{c|c|ccc}
learning rate & pretrain & AP & $\text{AP}_{50}$ & $\text{AP}_{75}$ \\ \hline \hline
0.45 \footnotesize(default) & SCRL & 40.9 & \textbf{62.5} & 44.5 \\
0.3 & SCRL & \textbf{41.2} & 62.4 & \textbf{45.1} \\ \hline
\end{tabular}
\vspace{1mm}
\caption{Performance improvement in COCO detection task with ResNet-50 trained for 1000 epochs, when using coarsely-tuned learning rate of 0.3.}
\label{tab:lr_search}
\end{table}

\subsection{Performance Improvement by Hyperparameter Search}
In all experiments in the paper, we naively transfer the sharable hyperparameters of BYOL to SCRL with which one can reproduce the performance reported in {\cite{BYOL}}. Although SCRL already outperforms BYOL under this condition, we observe an additional gain in downstream performance by tuning the learning rate alone with simple grid search, {\ie} +0.3 AP increase with the learning rate of 0.3 on COCO detection task, compared to the default learning rate, 0.45, as shown in Table {\ref{tab:lr_search}}.

\subsection{Using Negative Pairs for Upstream Training}
We exploit the negative pairs based on the SimCLR framework and obtain somewhat better results (+0.47 AP on COCO detection). However, this performance improvement requires an increased batch size and a sophisticated composition of the negative pairs, therefore we leave it for future works.

\subsection{Scale-invariant Representation Learning}
Before coming up with the feature-level matching, we had started from the baselines enforcing the input-level consistency. The best model share same details with BYOL but with the modification in the augmentation: spatially consistent cropping(or just use the entire image), and random aspect resizing followed by mean-padding to ensure the same spatial dimension for the sake of parallelized computation. 
We observe the performance degradation on the localization downstream task, in comparison to SCRL even with a single RoI pair. We hypothesize that this is due to the limitation in obtaining consistent local representation against an internal geometric translation of an object.

\end{document}